\documentstyle[ijcai89,named]{article}

\newcommand{\boldC}{\mbox{\bf C }}
\newcommand{\boldM}{\mbox{\bf M }}
\newcommand{\boldV}{\mbox{\bf V }}
\newcommand{\boldF}{\mbox{\bf F }}
\newcommand{\boldN}{\mbox{\bf N }}

\title{Context and ontology in understanding of dialogs}
\author{
Wlodek Zadrozny \\
\normalsize{IBM Research, T. J. Watson Research Center} \\
\normalsize{Yorktown Heights, NY  10598}  \\
\small{wlodz@watson.ibm.com}
\thanks{Proc. IJCAI'95 Workshop on Context in NLP. Montreal, Aug.1995}
}
\begin{document}
\maketitle

\begin{abstract}

We present a model of NLP in which ontology and context are
directly included in a grammar. The model is based on the
concept of {\em construction}, consisting of a set of
features of form, a set of semantic and pragmatic
conditions describing its application context,
and a description of its meaning.
In this model ontology is embedded into the grammar;
e.g. the hierarchy  of
{\it np} constructions is based on the corresponding ontology.
Ontology is also used in defining contextual parameters;
e.g. $\left[ current\_question \ time(\_) \right] $.

A parser based on this model allowed us to
build a set of dialog understanding systems that include
an on-line calendar, a banking machine, and an insurance quote
system. The proposed approach is an alternative
to the standard "pipeline"
design of morphology-syntax-semantics-pragmatics; the account of
meaning conforms to our intuitions about compositionality,
but there is no homomorphism from syntax to semantics.
\end{abstract}

\section{Introduction:
arguments for linking forms, meanings and contexts}

We present a new model of natural language based on the
concept of {\em construction}, consisting of a set of
features of form, a set of semantic and pragmatic
conditions describing its application context,
and a description of its meaning.
The model gives us a better
handle on phenomena of real language than
the standard approaches, such as syntax + semantics a la Montague.
It is also an alternative to the standard design based on the
pipeline syntax-semantics-pragmatics (we have little to say about
morphology at this point).
Since this work has been implemented, there is
also a computational argument in favor of this approach.

We claim that a linking of form, meaning and context
is needed to accurately describe NL constructions,
both "standard" and "non-standard". However,
we are not arguing for the unsuitability of syntax for
describing a "core" of language or a universal grammar.
We believe, that such a language
core is small, and that the syntactic descriptions of this
core are naturally paired with their meanings, which produces
a construction-based universal grammar. Furthermore,
new methods are needed
to handle phenomena of real languages. In this paper, our aim is not
to come with linguistic generalizations about universal
structural properties of sentences (although, of course, we have
nothing against them), but to come with an effective method for natural
language understanding, which in addition to computational effectiveness
would also have some linguistic and psychological plausibility. \\

The first argument for the linking of forms, meanings and contexts
is that it has been the experience of
of many people working in the field of computational linguistics
that combining syntactic and semantic information increases the
efficiency and effectiveness of processing; e.g.
\cite{Lytinen91} has shown that a semantically driven approach is
superior to syntax-first approach in processing text in narrow domains.

The second group of arguments is strictly linguistic.
Many linguistic phenomena can
be naturally described by the pairing of their syntactic
form with their semantic features. Such patterns of interdependence
of syntax and semantics are common both for standard constructions,
like NPs, VPs, or Ss, and for more exotic ones, such as sentences
with "let alone". Attempts to account for those phenomena in systems
without such interdependence of syntax and semantics lead to
drastic overgeneralizations.

As an example involving familiar constructions, consider
conjunctions. It is easily seen that
particular conjunctions select clauses with a specific logical
relation between their respective meanings:\\
\hspace*{.15in}  $^{\star}${\em This is a regular chair,
but you can sit on it.}\\
\hspace*{.15in} \ {\em  This is an old chair, but you can sit on it.}\\
In this case, the two phrases conjoined by "but" agree if
the second clause contradicts what is typically believed about entities
mentioned in the first clause. This kind of agreement
cannot be described without access to semantic information
and world knowledge.

Virtually every construction of English displays a similar
interplay of form and meaning.  We can refer the reader to
\cite{McCawley88}
for a discussion e.g. of the semantic significance of the
order of modifiers of
NPs, on the semantic conditions on the progressive, etc..
Regarding more exotic constructions,
Fillmore et al. observe that
standard grammars do not handle {\em open idioms}
such as
{\em The more you work, the easier it will get} or
{\em He doesn't eat fish, let alone shrimp.} They note that
each of these expressions "exhibits properties that are not fully
predictable from independently known properties of its
lexical make-up and its grammatical structure" \
\cite{Fillmal88} , p.\ 511.

To show that these types of
constructions (and the standard ones)
can be handled computationally,
\cite{Jurafsky92} proposed a construction formalism to
build "a linguistically motivated grammar (...) compatible
with psycholinguistic results on sentence processing".
(His grammar, however, does not include contextual information, nor
handle dialogs.)

Finally, we note that
context, especially dialog context, changes the range of applicable
constructions. Everybody knows that
{\it Kissinger thinks bananas} makes sense as an answer to
the question {\it What is (was) Nixon's favorite fruit}.
And, in spoken discourse context,
"fragments" behave like open idioms; e.g.
"afternoon" might mean "in the afternoon", "two" the second of the
choices (e.g. for a meeting time), and "checking" can stand
for "from (the) checking (account)", i.e. semantically be a "source". \\

The paper is organized as follows:
In Section 2, we present our NLU system and
an example that illustrates the need for contextual information
in understanding dialogs.
In Section 3 we describe
the formalism of constructions;
Section 4 presents examples of construction (from words to discourse).
Section 5 is about the role of  ontologies in the grammar.
Section 6 contains an example of parsing with a construction
grammar. In Section 7, we argue that construction-based approach
captures the intuition of
compositional semantics, and that it is an effective and reusable
encoding of linguistic knowledge.

\section{Grammar, context and ontology in a dialog system}

We have built a dialog understanding system, {\sc mincal},
for a small domain and a specific task, namely, for
scheduling (as well as canceling, and moving) meetings in an on-line
calendar \cite{Coling94}.
This work has later been extended
to other small discourse domains (fast food,
banking, insurance premium quotes etc.).

In {\sc mincal},
the grammar consists of about a hundred constructions and
a few hundred lexical entries,
and includes both sentential constructions and discourse constructions.
(Similar numbers for other applications).
The system also contains an encoding of elementary
facts about time and parameters of meetings; this background
knowledge is encoded in about a hundred Prolog clauses.

The system is capable of understanding dialogs such as this:\\
\hspace*{.15in} {\em --- Schedule a meeting with Bob. \\
\hspace*{.15in} --- At what time and date? \\
\hspace*{.15in} --- On August 30th. \\
\hspace*{.15in} --- At what time?     \\
\hspace*{.15in} --- 8.               \\
\hspace*{.15in} --- Morning or afternoon? \\
\hspace*{.15in} --- In the evening.} \\
It contains three main components with the following functions:
\begin{enumerate}
\item The {\it parser}
takes an input utterance from the user and a context
information from the discourse model.  It outputs the set of messages
corresponding to the possible meanings of
the uttered phrase.  The components of the parser are: the parser itself,
a lexicon, a set of constructions and a set of filters.
\item The {\it interpretation module (semantic engine)}
translates the messages it receives from the
parser to a set of slots with values, corresponding to the entities recognized
in the phrase,
like {\tt [action\_name reschedule],
[new\_event\_place "my manager's
office"] } etc.
\item The {\it discourse module}
gathers the computed slots, the current state
of its internal database and the current context.  From this it computes the
new context, changes the database if necessary, informs the user about the
action
taken, and carries on the dialog.
\end{enumerate}

{\sc mincal}
parser recognizes {\em Schedule a meeting with Bob}  as an instance
of {\em sent(imp)}, the imperative construction consisting of a verb
and an {\it np}, in this case {\em np(event)}.
\footnote{We will use the convention that {\it np, np(x), vp, vp(\_)}
etc. refer to elements of our construction grammar, while
traditional linguistic/syntactic categories will be denoted
by NP, VP, S etc.}
Then, in the context of the task and the question asked,
it parses, the subsequent sentences.
At each step of the dialog, the semantic engine extracts information
from the results of parsing to yield (at the end of the dialog):
\begin{verbatim}
***Slots:
[  [  action_name schedule]
   [  event_name  meeting]
   [  event_time
      [  [  minute 0] [ hour 20]
         [  day 30]   [ month 8]
   [  event_partner
      [  bob]
\end{verbatim}
At the end of the processing, another component of the
program changes the format of this data, and
schedules the meeting in an on-line calendar, {\it xdiary}.\\

We can make the  following observations.
(a) From the point of view of the operation of the system, we do not
care about the structure of the sentences (provided the parser
can handle them); we care only about their meanings,
and these meanings depend on context.
(b) The issue of ontology is not trivial.
Namely, we have
to distinguish not between three languages
(representations) and corresponding ontologies:
the language/ontology for parsing,
the language/ontology for representing information about the domain,
e.g. the calendar functions, and
the language/ontology for representing information about the
particular application, e.g. {\it xdiary}.
A natural question arising is the role of all three
representations in parsing.

We will talk about the interaction of form and meaning in the
next section. Now we want to make a few observations about
the context-dependence of interpreting dialogs.
In the first sentence of the above dialog,
the context is used to prevent
another reading in which {\em with Bob} modifies {\em schedule},
as in {\em Dance a tango with Bob!}. That is, we use a contextual
rule saying that for the calendar application people do not modify
actions, nor places.
By setting up a set of such domain- and application-specific filters,
we can remove ambiguities of PP attachment in most cases, e.g.\ in \\
\hspace*{.15in}  {\em Set up a lunch in the cafeteria with my boss} \\
\hspace*{.15in}  {\em Postpone the interview at 10 to Monday}. \\

In general, taking the context into account during parsing
allows the parser to focus on certain constructions rather than others,
as well as to compute certain meanings more accurately.
Using context to restrict the
constructions that are triggered during parsing
greatly reduces the number of edges our
parser has to consider, resulting in the increase of speed.
The dependence of meaning on context
is well known, e.g. deixis or deciding the reference of
pronouns. But
there are very natural examples not connected to deixis: in our dialog,
the expression {\em 8} is interpreted only as
a time expression (and not a place or something else),
because of the context of the preceding question
(asking about the time of the meeting).
Similarly,  for computing the meaning of other
time expressions: {\it 4 to 6} is more
likely to mean "beginning at 4 and ending at 6" than "5:56" in the
context of a question about the time of a meeting. But by using the
context the application, we can completely eliminate the second
reading, since all meetings are scheduled in 5 minute increments.\\

\section{Constructions: integrating forms, meanings and contexts}

\subsection{Constructions: The concept}

A {\em grammar} is a collection of constructions.
Each {\em construction} is given by the matrix:
\[
\left[
\begin{array}{l}
\boldN : name\_of\_construction \\
\left[
\begin{array}{lll}
\boldC & : & context \\
\boldV & : & structure / vehicle \\
\boldM & : & message / meaning \\
\end{array}
\right]
\end{array}
\right]
\]
The {\em structure / vehicle}
$\boldV$
consists of formulas describing presence (or perhaps
absence) of certain features of form within
the construction, e.g.
a list of subconstructions
and the way they have been put together -- in all our examples this is
concatenation, but there are
other possibilities, e.g.\ wrapping.
Since in practice generative grammars
use only taxemes that are "meaningful", i.e.\ one can associate some
meanings with their presence, we can include in our description of
constructions some of the features used by GB, GPSG or PEG
such as {\em  agreement} and lexical markings.
(cf. e.g.\ \cite{Sells85}
and \cite{Jensen86}).

The {\em context}
$\boldC$
consists of a set of semantic and pragmatic constraints limiting the
application of the construction. It can be viewed as a set of
{\em preconditions} that must be satisfied in order for a construction
to be used in parsing.
The {\em message} $\boldM$ describes the meaning
of the construction, via a set of syntactic,
semantic and pragmatic constraints.
\footnote{The language of meanings is
to a large extent an open problem, but
\cite{Talmy85} presents a catalogue of semantic relations
which are lexically expressible (in different languages). We believe
that an extension of this list can be used to classify meanings of all
constructions.}

For example, we can analyze expressions
{\it No, but I'll do it now/send it tomorrow/...}, typically given
as an answer to a question whether something has been done,
as a discourse construction.
Its description uses such relations as
{\em previous\_utterance/p\_utter},
{\em previous\_sentence/p\_sent} (i.e. the propositional content
of the previous utterance),
the message/meaning of $S$, $m(S)$.

\[
\left.
\begin{array}{l}
\boldN : sent(assrt, no.but.S) \\
\left[
\begin{array}{l}
       \boldC \; : \; \left[
           p\_utter(X) \& cons\_name(X, sent(ques,\_)) \right] \\
       \boldV \; : \; \left[
\begin{array}{l}
 struc("no"."but".S) \& \\
 cons\_name(S,sent(assrt,\_))
\end{array}
  \right] \\
       \boldM \; : \; \left[ p\_sent(Y) \& truth\_value(Y, 0) \&
                             m(S)  \right]
\end{array}
\right]
\end{array} \\
\right. \\
\]

As we can see, the construction applies only in the context of a
previously asked
question, and its message says that the answer to the question is negative,
after which it elaborates the answer with a sentence $S$. \\

There is no agreement on what is context, or, even more important,
on what is not. But, again, from the point of view of an abstract
computing device, a context is a collection of relations that are
interpreted as constraints. A partial list of such relations
can be found in the books on pragmatics
and it includes such relations as
{\em speaker},
{\em hearer},
{\em topic},
{\em presupposed},
{\em assertion, question, command, declaration, }
{\em ...     }. To this list we can add
{\em previous\_utterance,    }
{\em current\_domain,    }
{\em current\_application,    }
{\em current\_question,    }
{\em default\_value,     }
etc.

\subsection{Types and properties of constructions}

As we have said before, we see constructions as sets of
constraints on the relation between forms, meanings and contexts.
\cite{Nerbonne95}
argues that this view arises naturally in
computational linguistics, and discusses its implications for
the syntax-semantics interface and parsing.

We do not assume any specific constraint formalism; however
we do assume that the constraints can be inherited, and
they can be nonmonotonic, that is, more specific
constraints might invalidate the inherited ones.
Also, in the process of writing a grammar we do not have to specify
all constraints at once, e.g. for {\em no.but.S} construction
there should be a relationship, like shared topic, between
the sentence $S$ and the preceding question, however this constraint
should can be added at any time, and the lack of it should have
no impact on the rest of the grammar. We assume that the
difference of form implies the difference of meanings of two
constructions
(cf. \cite{Goldberg94} p.3),
but the opposite does not hold (ambiguity and multiple word senses).

Also we should note that apart from the inheritance hierarchy
grammars of constructions can be also partitioned "horizontally",
each class with its own subtheory.
For example,
\cite{Zwicky94}
divides constructions into four types:
\begin{enumerate}
\item {\em sentence-type} constructions;
\item {\em constituency} constructions, e.g. clauses, phrases, words;
\item {\em valency} constructions (head+dependents); and
\item {\em substitution} constructions (e.g. pro-forms and zeros).
\end{enumerate}

We can hypothesize the existence of a
{\em universal grammar of constructions}; e.g.
\mbox{}\cite{Goldberg94} \
p.5, says that
"simple clause constructions are associated directly with
structures which reflect scenes basic to human experience".
She then presents arguments supporting this thesis.
Similar arguments can be made for other types of constructions, e.g.
basic discourse constructions (cf. e.g.
\cite{Langacker91})
and anaphoric reference
\cite{Kay94}.

To end this general exposition, we list four basic differences between
construction grammars and other grammars.
(1) Construction grammars are not lexicalized; (2) They are not
head driven (since in real dialogs and texts it is often impossible
to find the head of a clause); (3) Parsing produces only flat, two
level, "semantic" structures -- which is important from the point of view
of efficiency; (4) Ontology plays an important role in organizing
the grammar and in guiding the parsing.

\section{Examples of constructions}

\subsection{Sentences and phrases}

The set of core constructions describes
an {\em agent} performing an {\em action}; the NP is as an
agent and a verb as an action. The English constructions
SV, SVO, SVOO, SVOC (subject-verb-object-complement), ...
could all be described in a similar fashion.
But we show only the SVO construction.

\[
\left[
\begin{array}{l}
\boldN : sent(assert,svo(\_)) \\
\left[
\begin{array}{l}
       \boldC \; : \; \left[ p2p, english \right] \\
       \boldF \; : \; \left[
\begin{array}{l}
 struc( NP1.VP.NP2) \& \\
 cons\_name(NP1,np(\_)) \& \\
 cons\_name(NP2,np(\_)) \& \\
 cons\_name(VP,vp(\_)) \& \\
\end{array}
  \right] \\
       \boldM \; : \; \left[
\begin{array}{l}
\left[ action , m(VP) \right] , \\
\left[ agent  , m(NP1) \right] , \\
\left[ object , m(NP2) \right] \\
\end{array}
                                     \right]
\end{array}
\right]
\end{array}
\right]
\]

The construction specifies that the action is given by the meaning
of the VP (which we assume would consist of a V and adverbials);
the agent and object are given by the meanings of the two NPs.\\

For computational reasons it is often convenient to assume that meanings
are expressed as lists of
$ \left[ attribute, value \right]$ pairs together
with formulas constraining those values. We put no
restrictions on either the form or the semantics of those formulas;
in particular they can refer to extralinguistic properties.
In the "atomic" case meanings consist simply of "denotations".
For example, for "dog" and "walk" we could have (respectively)
$ \left[ den, dog \right]$ and
$ \left[ den, walk \right]$. The SV construction would then
specify the meaning of the verb is combined with the meaning
of the subject (noun) to produce \
$$ \left[
  \left[ agent, dog \right] ,   \left[ action, walk \right]
\right]$$
Notice that at this point
the exact choice of notation is immaterial, e.g. we could
have written $den(dog)$ for $ \left[ den, dog \right] $.
The point is that constructions explicitly specify how
the meanings of subconstructions contribute to the meaning of
a whole, and this can be done in a number of ways:
by logical formulas, by combining features, functionally, etc.
We will use notation which seems the most easy to read.

Construction grammars make use of
defaults (that can be overridden).
Hence action verbs do not have to be
specified as such, but stative verbs must be specified as stative.
ACTIONS and verbs can be further subdivided, e.g. along the lines
proposed by
\cite{Dixon91}
into {\em motion-type, affect-type,
attention-type}, etc.. Correspondingly,  the defaults would be
elaborated into AGENT  is  MOVABLE for motion-type actions (walk,
run, etc.); OBJECT  is  TARGET and complement is
MANIP for {\em affect-type} actions (e.g.
kick); AGENT  is  PERCEIVER for {\em attention-type} verbs; and so on.\\

Similarly, nouns would form a hierarchy in which there is a general
category $n(\_)$, with a default meaning "entity",
and many subcategories $n(thing(\_))$,
$n(person(\_))$, $n(idea(\_))$,
$n(time(\_))$, $n(time(relative(\_)))$ (e.g. afternoon), etc.
(see Section 5 below).
The noun phrase can then be defined, as usual, as a noun with
modifiers ($MODS$), with the order to be specified for each
language separately.
And the default meaning of the NP is a conjunction
of the meaning of the noun with the relation defined by the modifiers.
Notice that the type of a noun phrase ($X$) is the type of its head
(as usual).

\[
\left[
\begin{array}{l}
\boldN : np(X) \\
\left[
\begin{array}{l}
       \boldC \; : \; \left[ nil \right] \\
       \boldF \; : \; \left[
\begin{array}{l}
 struc(\{ N , MODS \}) \& \\
 cons\_name({N},n(X)) \& \\
 cons\_name(MODS, T) \& \\
 noun\_modifier(T) \& \\
\end{array}
  \right] \\
       \boldM \; : \;  \left[ m(N) \ \& \
          m(MODS)(m(N))
                        \right]
\end{array}
\right]
\end{array}
\right]
\]

\subsection{Words as constructions}

We view languages as collections of constructions which range from
words to discourse. We have seen how the same representation
scheme can be used for different constructions.
In this subsection we apply it to lexical items. For instance,
the verbs "see" and "hit" can be represented as follows:

\[
\left[
\begin{array}{l}
\boldN : verb(perception(see)) \\
\left[
\begin{array}{lll}
\boldC & : & \left[
          english
        \right] \\
\boldV & : &  \left[ \begin{array}{l}
                    struc(see) \\
          \left[ subcat, \left[ subj(X), obj(Y) \right] \right]
             \end{array}
        \right]\\
\boldM & : &  \left[ \begin{array}{l}
          \left[
                perceiver, X  \right] ,
          \left[
                impression, Y \right] \\
             \end{array}
        \right]
\end{array}
\right]
\end{array}
\right]
\]

\[
\left[
\begin{array}{l}
\boldN : verb(affect(hit)) \\
\left[
\begin{array}{lll}
\boldC & : & \left[
          english
        \right] \\
\boldV & : &
              \left[ \begin{array}{l}
                     struc(hit) \\
          \left[ subcat, \left[ subj(X), \right. \right. \\
 \left. \left.     \ \ \ \ \ \ \ \ \ obj(Y), comp(Z)
           \right]
               \right]
             \end{array}
        \right] \\
\boldM & : &  \left[ \begin{array}{l}
          \left[ target, Y \right] ,
          \left[ manip , Z \right] \\
             \end{array}
        \right]
\end{array}
\right]
\end{array}
\right]
\]

The lexical entry specifies the semantic type of the verbs
(after
\cite{Dixon91}),
and the subcategorization information.
This information determines the default
semantics of the clause in which the verb appears.
Notice that (a)
even simple words require context to get an interpretation;
we say in $\boldC$ that the language code is {\em english} (but in other
cases it could also be spoken French, etc.); (b)
some pieces of information do not
have to be explicitly specified and can be
replaced by defaults.  E.g.
for "hit" we do not have to say that the subject is an agent;
it is enough to specify the semantic roles of the object and
the complement.

The simplicity of the meanings is a result of a deliberate
simplification.
In reality, the lexical meaning of any word is a much more
complicated matter.  Any review of issues of
of computational lexicography contains a list of the
types of features of lexical entries and inference methods
for natural language understanding.
For instance, in our lexicon
the messages of words may contain many of the attributes
that appear in the explanatory combinatorial dictionary of Melcuk (cf. e.g.
\cite{MelcukandPolguere87} and
\cite{Melcuk88}, pp.92-101) and the functions used there.

We end showing how one can write the matrices of the constructions
{\em pronoun(him)} and {\em determiner(the)}.

\[
\left[
\begin{array}{l}
\boldN : pronoun(him) \\
\left[
\begin{array}{lll}
\boldC & : & english \\
\boldV & : &
       \left[ \begin{array}{l}
             struc(him) \\
             \end{array}
        \right]\\
\boldM & : &
       \left[ \begin{array}{l}
          \left[ den,  he \right], \\
          \left[ case, acc \right]  \\
          \left[ person, third \right] \\
             \end{array}
        \right]\\
\end{array}
\right]
\end{array}
\right]
\]

\[
\left[
\begin{array}{l}
\boldN : determiner(the) \\
\left[
\begin{array}{lll}
\boldC & : &
       \left[ \begin{array}{l}
english, \\
     current\_discourse(DS) \ \& \\ DS \ne [nil] \
             \end{array}
\right] \\
\boldV & : &
       \left[ \begin{array}{l}
             struc(the) \\
          \left[ subcat, \left[ bare\_np(X) \right] \right],\\
             \end{array}
        \right] \\
\boldM & : &
       \left[ \begin{array}{l}
          \left[ definite,  m(X)  \right] \\
             \end{array}
        \right] \\
\end{array}
\right]
\end{array}
\right]
\]

Clearly, this is a very sketchy characterization of the definite
article. It says that there is a non-empty discourse context,
that it subcategorizes for a bare NP, and meaning is given
by the attribute {\em definite}. Although it captures
the main properties of "the",
we refer the reader to
\cite{Langacker91} p.96-103 and
\cite{Fauconnier85}
for a discussion of definite descriptions
and determiners, and for a characterization of
the attribute {\em definite} in terms of "mental spaces".

\section{Embedding ontology into grammar}

In our construction grammar
the representation of constructions is closely coupled
with a semantic taxonomy. Thus, for instance, not only do we
have an {\em np} construction, but also
such constructions as {\em np(place), np(duration),
np(time), np(time(hour))} etc. In other words, the semantic hierarchy
is reflected in the set of linguistic categories. (Notice that
categories are not the same as features).
Hence we do not have a list, but a tree of grammatical
categories.

To be more specific, let us consider temporal categories.
{\it Time} is divided into the following categories
{\it minute, hour, weekday, month,  relative,
         day\_part, week\_part, phase},
and more categories could be added if needed, e.g. {\it century}.
Some of these categories are further subdivided:
{\it day\_part} into {\it afternoon, morning, ...},
or
{\it relative(\_)} (e.g. {\it in two hours})
into {\it hour, minute, day\_part, week\_part, weekday}.\\

Note that (a) different constructions can describe the same
object of a given category, for example
{\it hour} can be given by a numeral or a numeral  followed by
the word "am" or "pm"; (b) there is a continuum of both categories
and constructions describing objects of a given category,
for instance, we have the following sequence of types of constructions
$$np > np(time) > np(time(month)) > $$
$$ >  np(time(month(january))) = january $$
corresponding to the sequence of categories
$$entity > time > month = time(month) > $$
$$ >  january =  time(month(january))  $$
In the first case {\it january} is a word (words are constructions,
too), in the second case it is a concept.\\

We end this sections with two notes. First, we can ask:
what about verbs and their hierarchies? --- At present
we have no hierarchy of actions, and no hierarchy of
verbs; perhaps when related
actions can appear as parameters of a plan such a need would arise.
However $np$ hierarchies help describe arguments of verbs.

Second, we would like to point to another, indirect, argument
in favor of the coupling of grammar and ontology. Namely,
the relationship between ontologies and case systems. For instance,
\mbox{}\cite{Copecketal92} list 28 cases grouped in 5 classes
(eg. {\it space (direction, orientation, location\_to, ...), or
causality (cause, contradiction, effect, ...)}).
Clearly, there is a relationship between cases and concepts in
ontological hierarchies.

\section{Parsing with constructions -- examples}

Parsing with constructions differs a bit from
syntactic parsing. First, the collection of features
that are used to drive parsing is richer, because it contains terms
with semantic and pragmatic interpretation. Second,
semantic and pragmatic information is used during parsing.
Third, descriptions assigned to strings by the parser
are different, namely,
the structural information is lost; instead the meanings/messages
are produced.

In the next example, notice that the knowledge
of application is needed to produce the following interpretation
of a complete sentence.
That is, the interpretation cannot be produced solely from
the edges of the chart produced by the parser; it must e.g. be known
that "set up" is means "schedule".
\begin{verbatim}
| ?- sem.
|: I want to set up an appointment on
   November 11.
[i,want,to,set,up,an,appointment,on,
 november,11]
***Slots:
[  [  action_name schedule]
   [  event_name meeting]
   [  event_time [ [  month 11]
                   [  day 11]
\end{verbatim}

Let us now consider the processing of fragments, which will also
explain how the interpretation above can be computed,
and how ontology is used in constructions. As noted before,
getting the slots from a fragment is possible only by
using context (such parameters as {\em current\_question} or
{\em topic}),
so that its meaning
can be unambiguously computed.
\footnote{In the following,
we do not show syntactic information, because syntactic features
are standard (e.g.  {\tt plural, 3rd, accusative, ...}).}
Let us assume that the system has asked for the time and date of
an appointment. The user however does not answer exactly to the point:
\begin{verbatim}
|: on November 11th with Martin.
[on,november,11,th,with,martin]
***Slots:
[  [  event_time    [ [ month 11]
                      [  day 11]
   [  event_partner [  martin]

| ?- li.
  Chart results: INACTIVE EDGES
* 0,1,[on] : prep(on) -> [on]
* 1,2,[november] : n(time(month)) -> [november]
  [[type,time(month)],[den,11]]
* 1,2,[november] :
  np(time(month)) -> [n(time(month))]
* 2,4,[11,th] :
  ordinal -> [numeral,st_nd_rd_th]
  [[den,11]]
\end{verbatim}
The preposition {\it on} by itself does not carry any meaning,
we treat it as a lexicalized feature.
In {\tt (2, 4)}, to prevent overgeneralizations such as
{\it 3 th}, the vehicle of the construction must contain the list
of cases {\it if 1 then st, if 2 then nd, ...}.
\begin{verbatim}
* 1,4,[november,11,th] :
  np(time(day)) -> [np(time(month)),ordinal]
  [[type,time],[den,[[month,11],[day,11]]]
* 0,4,[on,november,11,th] :
  pp(on,time) -> [prep(on),np(time(day))]
  [[type,event_time],[den,[[month,11],[day,11]]]
* 0,4,[on,november,11,th] :
  pp_list(on,time) -> [pp(on,time)]
* 0,6,[on,november,11,th,with,martin] :
  pp_list(with,person) ->
          [pp(on,time),pp_list(with,person)]
  [[pp_msg,[[type,partner],[den,martin]]],
   [pp_msg,[[type,event_time],
            [den,[[month,11],[day,11]]]]
* 0,6,[on,november,11,th,with,martin] :
  pp_list(on,time) ->
            [pp_list(on,time),pp(with,person)]
\end{verbatim}

The meaning of the {\it pp} {\tt (0, 4)}  is a simple transformation of
the meaning of the {\it np} in  {\tt (1, 4)}. But notice that this
happens because we have a specific construction which says that
a combination of {\it on} with an {\it np(time(\_))} produces
{\it event\_time}. Altogether, we have encoded a few dozens
various {\it pp} constructions, but in a given application
only a fraction of them are used.

In the same context, please note that, because of the close
relationship between the domain ontology and the hierarchy of
constructions, we can also postulate a close relationship
between the type of meaning a construction expresses and
its category (i.e. name), for example, the type of meaning of
{\it np(time(hour))} is {\tt [type  event\_time]}.

At the end, we obtain two parses of {\tt (0, 6)}, but they
have the same messages. The difference lies in the category assigned
to {\it pp\_list}; and in this particular case the choice of
that category is not important.
\footnote{
We have experimented with different representations for constructions,
and we have built two parsers of constructions, of which one
was a postprocessor to
a standard grammar, the English Slot Grammar (ESG) of McCord,
and it is described in \cite{Zad92ait}.}

\section{Knowledge and meaning}

\subsection{Compositionality}

The idea of {\em compositionality} is used
to account for the ability of the language user to understand
the meaning of sentences not encountered before. The new sentence
can be understood, because it is composed of parts (words) that
the user knows, and the meaning of the new sentence is computable
from the meanings of the words that compose it. But the standard
definition of compositionality is formally vacuous
\cite{lp95} (i.e. any semantics can be compositional), so
the concept of compositionality has to be reexamined.

Given that, note that a grammar of constructions can account
for the ability of the language user to understand
the meaning of novel sentences without separating the language into
syntax, semantics and pragmatics. Namely, the meaning of a larger
construction is a combination of the meanings of its parts
(and the way they are put together). The message of the larger
construction specifies how its meaning depends on the meanings
of the parts.

\subsection{Ontology and knowledge bases}

We view construction grammars as representations of domain-independent
linguistic knowledge. That is why we need a domain-dependent
semantic module to map linguistic representations into concepts of
the domain. The fact that we have been able to use the same
sets of constructions for various domains is an argument for
feasibility of this approach.

Obviously NLU is impossible without access to vast bodies of
background knowledge. The fact that meanings of NL utterances
are classified and described with the help of general ontologies
suggests that linking linguistic and non-linguistic
knowledge for the purpose of reasoning about dialogs and texts
might be possible without the help of a translator program.
Furthermore, it might be possible to modularize those sources
of knowledge.

\section{Conclusions}

The innovations we have proposed ---
the close coupling of semantic hierarchies with linguistic
categories, the use of
context in representing linguistic knowledge,
and the representation of the grammar as a dictionary of
constructions --- not only facilitate the development of the
grammar, but also its interaction with the inference engine
and the application.

Obviously, the usefulness
of this approach is not limited to natural language interfaces.
We have used parts of the grammar of constructions
for some information retrieval tasks; and we can easily
imagine it being applied to
text skimming and to machine translation in limited domains.

The approach to language understanding we
are advocating has several advantages: it
agrees with the facts of language; it is not restricted to
a "core" of language, but applies to standard and more exotic
constructions, including language fragments;
and it is computationally feasible, as it has been
implemented in a complete working system.\\

Summarizing,
the most important innovations implemented in the system are
the close coupling of semantic hierarchies with the set of linguistic
categories; the use of
context in representing linguistic knowledge, esp. for discourse
constructions; and a non-lexicalist encoding of the grammar
in a dictionary of constructions.
We have obtained a new language model in which forms cannot
be separated from meanings. We can talk about meaning in
a systematic way, but we do not have compositionality
described as a homomorphism from syntax to semantics.
(This is theoretically interesting, also because
it is closer to intuitions than current models of semantics).
We have validated this model by building
prototypes of natural language interfaces.


\end{document}